# MFCC BASED ENLARGEMENT OF THE TRAINING SET FOR EMOTION RECOGNITION IN SPEECH

Inma Mohino-Herranz[1], Roberto Gil-Pita[1], Sagrario Alonso-Diaz[2] and
Manuel Rosa-Zurera[1]

[1]Department of Signal Theory and Communications, University of Alcala, Spain
[2]Human Factors Unit, Technological Institute "La Marañosa" –Ministry of Defense,
Madrid (Spain)

## ABSTRACT

*Emotional state recognition through speech is being a very interesting research topic nowadays. Using subliminal information of speech, denominated as "prosody", it is possible to recognize the emotional state of the person. One of the main problems in the design of automatic emotion recognition systems is the small number of available patterns. This fact makes the learning process more difficult, due to the generalization problems that arise under these conditions.*

*In this work we propose a solution to this problem consisting in enlarging the training set through the creation the new virtual patterns. In the case of emotional speech, most of the emotional information is included in speed and pitch variations. So, a change in the average pitch that does not modify neither the speed nor the pitch variations does not affect the expressed emotion. Thus, we use this prior information in order to create new patterns applying a gender dependent pitch shift modification in the feature extraction process of the classification system. For this purpose, we propose a frequency scaling modification of the Mel Frequency Cepstral Coefficients, used to classify the emotion. For this purpose, we propose a gender dependent frequency scaling modification. This proposed process allows us to synthetically increase the number of available patterns in the training set, thus increasing the generalization capability of the system and reducing the test error. Results carried out with two different classifiers with different degree of generalization capability demonstrate the suitability of the proposal.*

## 1. INTRODUCTION

Emotional state recognition (ESR) through speech is being a very interesting research topic nowadays. Using subliminal information of speech, it is possible to recognize the emotional state of the person. This information, denominated "prosody", reflects some features of the speaker and adds information to the communication [1], [2].

The standard scheme of an ESR system consists of a feature extraction stage followed by a classification stage. Some of the most useful features used in speech-based ESR systems are the Mel-Frequency Cepstral Coefficients (MFCCs), which are one of the most powerful features used in speech information retrieval [3]. The classification stage uses artificial intelligence techniques to learn from data in order to determine the classification rule. It is important to highlight that in order to avoid loss of generalization of the results, it is also necessary to split the available data in two sets, one for training the system and other for testing it, since the data must be different in order to avoid loss of generalization of the results.

One of the main problems in the design of automatic ESR systems is the small number of available patterns. This fact makes the learning process more difficult, due to the generalization problems that arise under these conditions [4], [5].

A possible solution to this problem consists in enlarging the training set through the creation the new virtual patterns. This idea, originally proposed in [6], consists in the use of auxiliary information, denominated hints, about the target function to guide the learning process. The use of hints has been proposed several times in several applications, like, for instance, automatic target recognition [7], or face recognition [8].

In the case of emotional speech, it is important to highlight that most of the information is included in speed and pitch variations [9]. On the other hand, the average pitch of the sentence is mainly dependent on the individual. Since the emotion recognition task in the problem at hand must be carried out independently on the specific characteristics of a given subject, a change in the average pitch value that does not modify neither the speed nor the pitch variations does not affect the expressed emotion. It is important to highlight that in the particular case of ESR systems tailored for a specific individual the proposed solution could not be suitable, since in that case a change in the global pitch could represent a variation in the emotion instead of a variation of the main characteristics of the subject.

In this work we propose the creation of new patterns by applying a pitch shift modification in the feature extraction process of a multi-subject ESR system. For every pattern in the training set, we apply a set of pitch shifts through frequency scaling in the MFCC extraction process. So, several new virtual patterns are generated from each pattern in the training set using a range of shifts for the pitch. The size, density and shape of the range of the applied pitch shifts are parameters of the proposed enlargement method, and their effect over the final results is also studied. Furthermore, since the gender is related to the range of possible valid pitches [10], [11], we propose to use it in order to modulate the shape of the range of pitch variation, avoiding the creation of non-valid pitches. This proposed process allows us to synthetically increase the number of available patterns in the training set, thus increasing the generalization capability of the system and reducing the test error.

In order to demonstrate the suitability of the proposal, two different classifiers (the Least Square Linear Classifier and the Least Square Diagonal Quadratic Classifier) have been tested under a set of experiments using an available database. These classifiers have different generalization capabilities and serve to demonstrate the performance of the proposed enlargement under different scenarios of generalization problems.

## 2. MATERIALS AND METHODS

This section explains the two main stages of an ESR system: the feature extraction stage and the classification stage, describing the configuration of the ESR system used in the experiments.

### 2.1. Feature extraction: Mel-Frequency Cepstral Coefficients (MFCCs)

Obtaining MFCC coefficients [12] has been regarded as one of the techniques of parameterization most important used in speech processing. They provide a compact representation of the spectral envelope, so that most of the energy is concentrated in the first coefficients. Perceptual analysis emulates human ear non-linear frequency response by creating a set of filters on non-linearly spaced frequency bands. Mel cepstral analysis uses the Mel scale and a cepstral smoothing in order to get the final smoothed spectrum. Figure 1 shows the scheme for the MFCC evaluation.

The main stages of MFCC analysis are:

- *Windowed*: In order to overcome the non-stationary of speech, it is necessary to analyse the signal in short time periods, in which it can be considered almost stationary. So, time frames or segments are obtained dividing the signal. This process is called windowed. In order to maintain continuity of information signal, it is common to perform the windowed sample with frame blocks overlap one another, so that the information is not lost in the transition between windows.

- *DFT*: Following the windowed, DFT is calculated to $x_t[n]$, the result of windowing the *t*-th time frame with a window of length $N$.

$$X_t[k] = \sum_{n=0}^{N-1} x_t[n] \cdot e^{-j2\pi nk}, 0 \leq k \leq N-1 \tag{1}$$

From this moment, phase is discarded and we work with the energy of speech signal, $|X_t[k]|^2$.

- *Filter bank*: The signal $|X_t[k]|^2$ is then multiplied by a triangular filter bank, using Equation (2).

$$E_{mt} = \sum_{k=0}^{N/2} |X_t[k]|^2 H_m[k], \quad 1 \leq m \leq F \tag{2}$$

where $H_m[k]$ are the triangular filter responses, whose area is unity. These triangles are spaced according to the MEL frequency scale. The bandwidth of the triangular filters is determined by the distribution of the central frequency $f[m]$, which is function of the sampling frequency and the number of filters. If the number of filters is increased, the bandwidth is reduced.

So, in order to determine the central frequencies of the filters $f[m]$, the behaviour of the human psychoacoustic system is approximated through $B(f)$, the frequency in MEL scale, in Equation (3).

$$B(f) = 2595 \cdot log(1 + f/700) \tag{3}$$

where $f$ corresponds with the frequency represented on a linear scale axis.

Therefore, the triangular filters can be expressed using Equation (4).

$$H_m[k] = \begin{cases} 0, & k < f[m-1] \\[2mm] \frac{2(k-f[m-1])}{(f[m+1]-f[m-1])(f[m]-f[m-1])}, & f[m-1] \leq k < f[m] \\[2mm] \frac{2(f[m+1]-k)}{(f[m+1]-f[m-1])(f[m]-f[m-1])}, & f[m] \leq k < f[m+1] \\[2mm] 0 & k \geq f[m+1] \end{cases} \tag{4}$$

where $1 \leq m \leq F$ , being $N$ the number of filter, and furthermore we have the central frequency $f[m]$ of the *m*-th frequency band :

$$f[m] = \frac{N}{F_s} B^{-1} \left( m \frac{B(F_s/2)}{F+1} \right) \tag{5}$$

where $B^{-1}(b) = 700(e^{b/2595} - 1)$, and $F_s$ is the frequency sampling.

- *DCT (Discrete Cosine Transform)*: Through the DCT, expressed in Equation (6), the spectral coefficients are trans- formed to the frequency domain, so the spectral coefficients are converted to cepstral coefficients.

$$MFCC_{mt} = \sum_{k=1}^{F} log(E_{mk}) cos(m(k-1/2)\pi/N), \quad m = 1, \cdots, F$$

(6)

The MFCCs are evaluated, features are determined from statistics of each MFCC. Some of the most common used statistics are the mean and the standard deviation. In is also habitual to use statistics from differential values of the MFCCs, denominated, delta MFCC, or ΔMFCCs. These ΔMFCCs are determined using Equation (7),

$$\Delta MFCC_{mt} = MFCC_{mt} - MFCC_{m(t-d)}$$

(7)

where $d$ determines the differentiation shift. In this paper we use as features the mean and standard deviation of the MFCCs, and the standard deviation of ΔMFCCs with $d = 2$, since we have found that these values obtain very good results with a considerably low number of features.

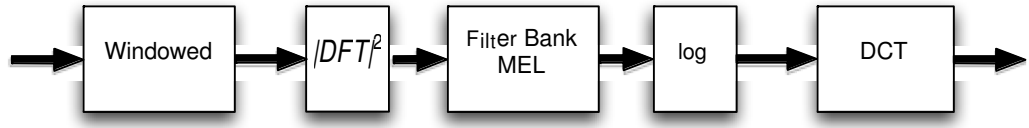

Figure 1 Scheme to MFCC calculate

## 2.2. Classification stage

This section aims to explain the classifier used, the Least Square Linear Classifier and the Least Square Diagonal Quadratic Classifier.

### 2.2.1. Least Square Linear Classifier

Let us define a set of training patterns $x=[x_1, x_2, \ldots, x_L]^T$, where each of these patterns corresponds to one of the possible classes denoted as $C_i$, $i=1, \ldots, K$. In a linear classifier, the decision rule is obtained using a set of $K$ linear combinations of the training patterns, as shown in Equation (8).

$$y_k = w_{k0} + \sum_{n=1}^{L} w_{kn} x_n$$

(8)

The design of the classifier consists in finding the best values of $w_{kn}$ and $w_k$ in order to minimize the classification error.

The output of the linear combinations $y_k$ is used to determine the decision rule. For instance, if the component $y_k$ gives the maximum value of the vector, then the $k$-th class is assigned to the pattern. In order to determine the values of the weights, it is necessary to minimize the mean squared error value, which can be carried out using the Wiener Hopf equations [13].

This classifier is very simple, because the boundaries are hyperplanes. Thus, and due to the simplicity of the implemented decision boundaries, both the error performance and the generalization capability use to be high.

### 2.2.2. Least Square Diagonal Quadratic Classifier

The Least Square Diagonal Quadratic Classifier is a classifier that renders very good results with a very fast learning process [14] and therefore it has been selected for the experiments carried out in this paper. Let us considerer a set of training patterns $x = [x_1, x_2, \ldots, x_L]^T$, where each of these patterns is assigned to one of the possible classes denoted as $C_i$, $i = 1, \ldots, k$. In a quadratic classifier, the decision rule can be obtaining using a set of $k$ combinations, as shows Equation (9).

$$y_k = w_{k0} + \sum_{n=1}^{L} w_{kn} x_n + \sum_{n=1}^{L} \sum_{m=1}^{n} x_m x_n v_{mnk} \qquad (9)$$

where $w_{kn}$ and $v_{mnk}$ are the linear and quadratic values weighting respectively.

This classifier is similar to the one above, but the boundaries are quadratic functions, which implies that the complexity of the system is more elevated. Implies that the intelligence of the classifier allows us to obtain better results in the study of this error probability. However, this classifier presents generalization problems higher than the Linear Classifier.

## 3. PROPOSED MFCC-BASED ENLARGEMENT OF THE TRAINING SET

As we stated in the introduction, it is important to highlight that most of the information of emotional speech is included in speed and the pitch variations [14]. So, an average change in the pitch value that does not modify neither the speed nor the pitch variations does not affect the expressed emotion.

In this paper we propose to modify the MFCC extraction in order to implement frequency scaling, allowing to create new patterns for the training set. So, the MFCCs can be easily pitch-shifted through a scale factor applied in frequency domain. This modification is applied to each pattern in the database, allowing to enlarge the training set.

Let us define the Pitch Shift Factor ($P_{SF}$) as a global change of the pitch, measured in semitones. Then, this shift in the pitch is equivalent to scaling the frequency with a Frequency Scale Factor ($F_{SF}$). So, the relationship between $P_{SF}$ and $F_{SF}$ can be expressed using Equation (10).

$$F_{SF} = 2^{\frac{P_{SF}}{12}} \qquad (10)$$

In order to apply this frequency scaling in the MFCC process, the central frequencies $f[m]$ of the triangular filters are modified, taking into account the scaled frequency factor. So, in Equation (11) we can observe the relationship between the original and synthetic frequency.

$$f'[m] = F_{SF} \cdot f[m] \qquad (11)$$

Being the new frequency scale, as shows in Equation (12)

$$f'[m] = F_{SF} \cdot \frac{N}{F_s} B^{-1} \left( m \frac{B(F_s/2)}{M+1} \right) \qquad (12)$$

In the Figure 2, we can observe the difference between the standard relationships between center frequency for each coefficient, and the difference when the frequency has been scaled. So, the center frequency is reduced when increase the number of cepstral coefficients.

As an example, the difference between the MFCCs calculated with $P_{SF}=0$ and $P_{SF}=1$ are shown in Figure 3. So, we can observe the filter responses in logarithmic scale without frequency shift of MFCCs with a shifting in frequency of one semitone.

As mentioned, we propose the creation of new patterns by applying a pitch shift modification in the feature extraction process of a multi-subject ESR system. For every pattern in the training set, we apply a set of pitch shifts through frequency scaling in the MFCC extraction process. So, several new virtual patterns are generated from each pattern in the training set using a range of shifts for the pitch. Furthermore, since the gender is related to the range of possible valid pitches, we have used this information in order to modulate the shape of the range of pitch variations, avoiding the creation of non-valid pitches.

In order to implement the enlargement of the database using pitch shifting, three factors must be taken into account: the range of the pitch shifting ($R$) and the step of the pitch shifting ($S$) and the symmetry factor (K).

- *Range (R):* The range defines the maximum absolute variation in the pitch modification process in semitones. With this parameter it is possible to change the upper and lower limits of the shift variations.

- *Step (S):* The step defines the smallest change in the pitch that is produced in the pitch shifting process in semitones.

- *Symmetry factor (K):* This factor controls the symmetry factor of the range, that is, the relationship between the maximum positive variation of the pitch and the minimum negative variation of the pitch. The key point is that those files corresponding to a male speaker are mainly positively shifted, and those files with a female speaker are mainly negatively shifted. So, K modulates the minimum value of the range of variation for males and the maximum value of the range of variation for females. Thus, it is possible to define the range shift to $P_{SF} \left[ -K \cdot R, R \right]$ for the case of males. In the case of females, the range shift used for calculating the $P_{SF}$ is $\left[ -R, K \cdot R \right]$. For instance, R=2, S=0.5 and K=0.75, this implies that the range to the males is $\left[ -1.5, 2 \right]$ and in the case of females is $\left[ 2, 1.5 \right]$.

Taking into account these three factors, it is possible to determine the enlargement factor (*EF*), that is, the number of times that the size of the training set is increased.

$$EF = \left\lfloor \frac{R(1+K)}{S} \right\rfloor + 1 \qquad (13)$$

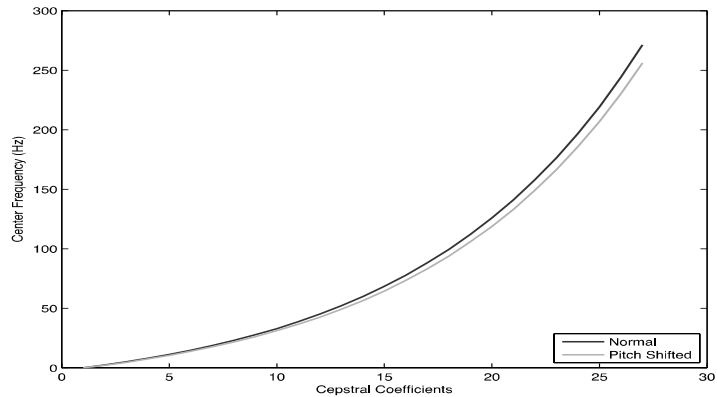

Figure 2 Central frequency *f[m]* of the triangular filter, original and modified

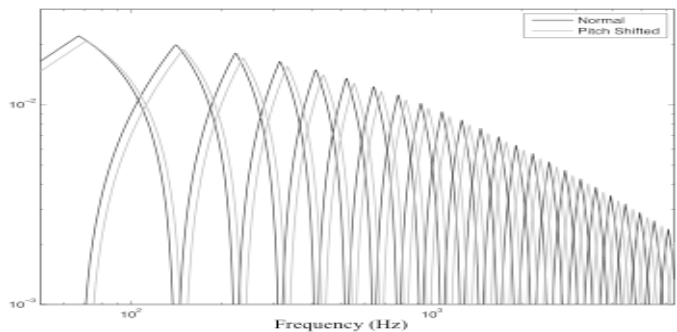

Figure 3 Triangular filters *H[m]* for the calculate of the MFCC, original and modified

34

# 4. RESULTS

This section shows the different experiments done and the results to highlight.

In this study, we have used the public database "The Berlin Database of Emotional Speech" [15]. This database consists of 800 files of 10 actors (5 males and 5 females), where each actor produces 10 German utterances (5 short and 5 longer sentences) simulating seven different emotions. These emotions are: Neutral, Anger, Fear, Happiness, Sadness, Disgust, and Boredom. The recordings were using a sampling frequency of 48 kHz and later downsampled to 16 kHz. Although this database consists of 800 files, 265 were eliminated, since only those utterances with a recognition rate better than 80% and naturalness better than 60% were finally chosen. So, the database consists of 535 files.

Since the size of the database is not very large, and in order to evaluate the results and to ensure that they are independent of the partition between training set and test set, we have used the validation method denominated Leave One Out [16] [17]. This is a model validation technique to evaluate how the results of a statistical analysis generalize to an independent data set. This method is used in environments where the main goal is the prediction and we want to estimate how accurate is a model that will be implemented in practice.

This technique basically consists in tree stages:

- First, the database is divided into complementary subsets called: training set and test set, where the test sets contains only one pattern in the database.

- Then, the parameters of the classification system are obtained using the training set.

- Finally, the performance of the classification system is obtained using the test set.

In order to increase the accuracy of the error estimation while maximizing the size of the training set, multiple iteration of this process are performed using a different partitions each time, and the test results are averaged over the different iterations.

In this paper we use an adaptation of this technique to the problem at hand, which we denominate *Leave One Couple Out*. So, we have worked with the database discussed above, which consists of 5 male and 5 female. In this case, we used 4 male and 4 female for each training set and 1 male and 1 female for each test set. This division guarantees complete independence between training and test data, keeping a balance in the gender. Therefore, our leave one couple out is repeated 25 times, using each iteration different training and test sets.

Concerning the features, a window size of N = 512 has been used, which implies time frames of 32ms. We have then selected mean and standard deviation of 25 MFCCs, and standard deviation of 2-ΔMFCCS, resulting in a total of 75 features, which has been used to design a linear and quadratic classifier. In order to complete the comprehension of the results obtained, it is necessary to analyse the error probability for training set, the error probability for test set and the enlargement factor (EF).

This work aims at the enlargement of the database available. For this purpose we have modified the average pitch in the available patterns through the frequency scaling the MFCCs. It is important to highlight that, in order to make the results comparable, the test set has not been enlarged nor modified in the experiments carried out in this paper.

In order to demonstrate the results obtained, we have done different experiments with different parameters. The main parameters varied in the experiments are: Range (*R*), Step (*S*) and Symmetry Coefficient (*K*). The experiments have been done with a wide range of values for each parameters used. Hence, the values used in order to calculate the Range (*R*) are 0, 0.5, 1, 2, 3, 4, 5, 6, 8, 10 y 12. These values have been chosen in order to clearly show the evolution of the results. For the second parameter the Step (*S*), we have taken values from 1/32 to 2, being the values chosen, 1/32, 1/16, 1/8, 1/4, 1/2, 1 and 2.

At last, the values for the Symmetry coefficient, are only two, $K = 1$ and $K = 0.75$. We have checked with others different values of $K$, but using these values it is possible to see the response of the system under different conditions.

The Figure 4 shows the relationship between the Error Probability for the test set and the Enlargement Factor using the Linear Classifier. It is possible to observe that the results are very similar for both values of symmetry coefficient, that is, $K=1$ and $K=0.75$, even the error probability tends to slightly increase. The minimum error probability for the test set is around 30.80% using $K=0.75$. This percentage is very similar to the result obtained using the $K=1$. In the Figure 4 is possible to observe the different values of error probability in function of the $EF$. Being, the Error Probability around 33.7% with an $EF=1$. However, using $EF=29$ we can to obtain the Error Probability around 30.8%. It implies, using the enlargement of the training set for emotion recognition in speech it is possible to reduce the error probability for the test set.

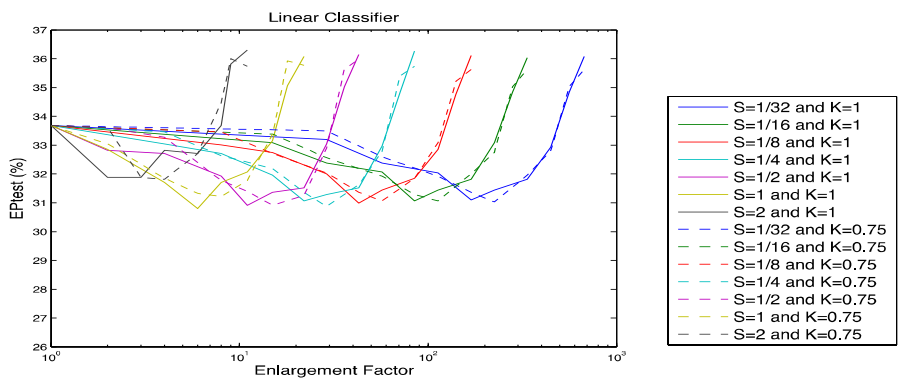

Figure 4 Relationship between the Error Probability for the test set and the Enlargement Factor. Linear Classifier

However, if we study the performance of the system using the Quadratic Classifier as shows the Figure 5, the error probability for the test set decreases around 3% with respect to the Linear Classifier. Additionally, the difference between the results obtained to the different symmetry coefficients shows the importance of this coefficient, since the error probability is reduced almost every value of $S$.

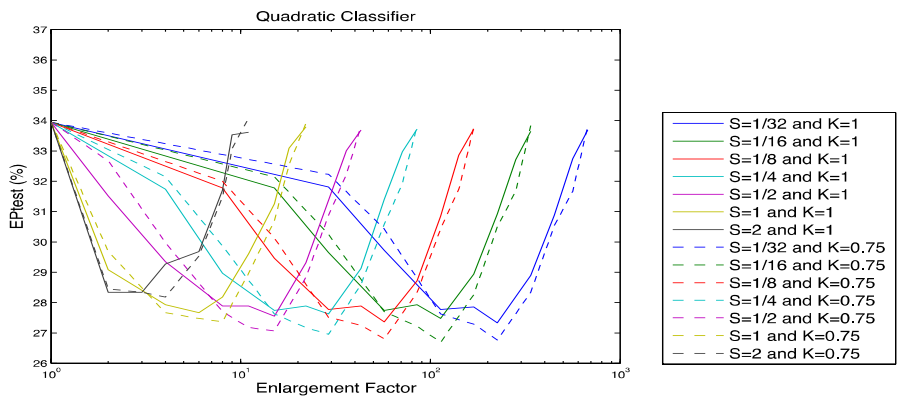

Figure 5 Relationship between the Error Probability for the test set and the Enlargement Factor. Quadratic Classifier

As we have demonstrated in the Figures 4 and 5, the Error Probability for the test set improve with $K$=0.75.

In order to study the response of the system with different values of $K$, we have check several values, but the best results correspond to the $K$=0.75, hence it has been selected to show in the next tables.

The Figure 6 it is possible to observe the evolution and trend of the error probability for the test set and training set in different cases, such as, using the Linear Classifier and the Quadratic Classifier. The error probability for the test set is around 30%. Besides, it is possible to see that using Range $R$= 4 the error probability for the test set is lower than other $R$.

However, if we pay attention to the trend of the error probability for the training set, it is easy to observe a continue increase of the error probability with respect to the Range ($R$).

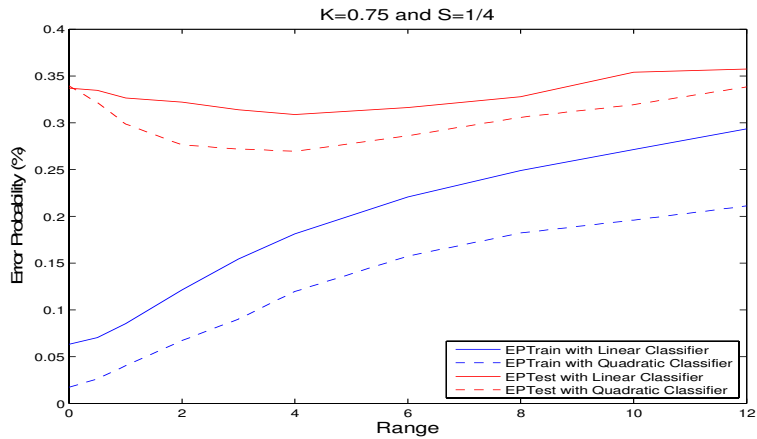

Figure 6 Linear Classifier vs. Quadratic Classifier. K = 0.75 and S=1/4

The Table 1 shows the Error Probability for the test set using the Linear Classifier for 10 different values of Range ($R$) and 7 different values of Step ($S$). The symmetry coefficient used is $K$=0.75, since it is demonstrated above that it provides better results. In this table, it is possible to observe the trend of the error probability with $R$. Furthermore, the $S$ that provides better results is $S$=1/4. Being the error probability less around 30%. In order to get to this results is necessary an enlargement factor ($EF$) of 29. That is, it is necessary increase 29 times the patterns set. In this Table it is possible to compare the results obtained when de $EF$=1, being 33.68% and when the $EF$ is increased until 29, we obtain an error probability for the test set of 30.88%. It implies the improved performance with enlargement for the training set.

The EF in each case is shown in the Table 3.

Table 1 Error Probability for the test set using the Linear Classifier

| Error Probability | | Range | | | | | | | | | |
|---|---|---|---|---|---|---|---|---|---|---|---|
| | | 0 | 0.5 | 1 | 2 | 3 | 4 | 6 | 8 | 10 | 12 |
| Step | 1/32 | 33.68% | 33.50% | 32.60% | 31.93% | 31.36% | 31.03% | 31.96% | 32.82% | 35.03% | 35.63% |
| | 1/16 | 33.68% | 33.38% | 32.56% | 31.93% | 31.29% | 31.07% | 32.00% | 32.75% | 35.03% | 35.63% |
| | 1/8 | 33.68% | 33.46% | 32.71% | 32.04% | 31.36% | 31.07% | 31.85% | 33.08% | 35.21% | 35.63% |
| | 1/4 | 33.68% | 33.46% | 32.64% | 32.19% | 31.36% | **30.88%** | 31.63% | 32.79% | 35.40% | 35.74% |
| | 1/2 | 33.68% | 33.61% | 33.27% | 31.78% | 31.40% | 30.92% | 31.25% | 32.90% | 35.63% | 36.00% |
| | 1 | 33.68% | 33.68% | 33.05% | 31.85% | 31.33% | 31.21% | 31.78% | 33.38% | 35.93% | 35.78% |
| | 2 | 33.68% | 33.68% | 33.68% | 33.61% | 31.89% | 31.81% | 32.71% | 34.47% | 36.00% | 35.74% |

In the Table 2, it is possible to observe the Error Probability for the test set using the Quadratic Classifier. The Error Probabilities are less than those obtained in the Table 1 using the Linear Classifier. In the case of the Quadratic Classifier, it is possible to reduce the Error Probability to 26.69%. This result is obtained with $R$=4, $S$=1/16 and $EF$= 113. The Enlargement Factor used is shown in the Table 3. However, if $EF$=1 the Error probability for the test set is 33.94%. It demonstrates that when the $EF$ is increased until certain values it is possible to improve the results obtained.

Table 2 Error Probability for the test set using the Quadratic Classifier

| Error Probability | | Range | | | | | | | | | |
|---|---|---|---|---|---|---|---|---|---|---|---|
| | | 0 | 0.5 | 1 | 2 | 3 | 4 | 6 | 8 | 10 | 12 |
| Step | 1/32 | 33.94% | 32.22% | 30.43% | 27.63% | 27.29% | 26.77% | 28.34% | 30.54% | 31.66% | 33.79% |
| | 1/16 | 33.94% | 32.15% | 30.24% | 27.70% | 27.21% | **26.69%** | 28.26% | 30.50% | 31.66% | 33.83% |
| | 1/8 | 33.94% | 32.00% | 30.13% | 27.51% | 27.25% | 26.80% | 28.34% | 30.43% | 31.74% | 33.76% |
| | 1/4 | 33.94% | 32.15% | 29.87% | 27.63% | 27.18% | 26.95% | 28.60% | 30.58% | 31.93% | 33.83% |
| | 1/2 | 33.94% | 32.67% | 29.98% | 27.70% | 27.18% | 27.07% | 28.82% | 30.88% | 32.22% | 33.83% |
| | 1 | 33.94% | 33.94% | 29.68% | 27.66% | 27.48% | 27.36% | 28.97% | 30.69% | 32.56% | 33.91% |
| | 2 | 33.94% | 33.94% | 33.94% | 28.45% | 28.34% | 28.19% | 29.53% | 31.48% | 33.08% | 34.09% |

Table 3 Enlargement Factor

| Enlargement Factor | | Range | | | | | | | | | |
|---|---|---|---|---|---|---|---|---|---|---|---|
| | | 0 | 0.5 | 1 | 2 | 3 | 4 | 6 | 8 | 10 | 12 |
| Step | 1/32 | 1 | 29 | 57 | 113 | 169 | 225 | 337 | 449 | 561 | 673 |
| | 1/16 | 1 | 15 | 29 | 57 | 85 | 113 | 169 | 225 | 281 | 337 |
| | 1/8 | 1 | 8 | 15 | 29 | 43 | 57 | 85 | 113 | 141 | 169 |
| | 1/4 | 1 | 4 | 8 | 15 | 22 | 29 | 43 | 57 | 71 | 85 |
| | 1/2 | 1 | 2 | 4 | 8 | 11 | 15 | 22 | 29 | 36 | 43 |
| | 1 | 1 | 1 | 2 | 4 | 6 | 8 | 11 | 15 | 18 | 22 |
| | 2 | 1 | 1 | 1 | 2 | 3 | 4 | 6 | 8 | 9 | 11 |

## 5. CONCLUSIONS

In the study of emotions in speech, one of the main problems is the small number of available patterns. This fact makes the learning process more difficult, due to the generalization problems in the learning stage. In this work we propose a solution to this problem consisting in enlarging the training set through the creation the new virtual patterns. In the case of emotional speech, most of the emotional information is included in speed and pitch variations. Thus, a change in the average pitch value that does not modify neither the speed nor the pitch variations does not affect the expressed emotion. So, we use this prior information in order to create new patterns applying a pitch shift modification in the feature extraction process of the classification system. For this purpose, we propose a gender dependent frequency scaling modification. This proposed process allows us to synthetically increase the number of available patterns in the training set, thus increasing the generalization capability of the system and reducing the test error.

In order to demonstrate the suitability of the proposal, two different classifiers (the Least Square Linear Classifier and the Least Square Diagonal Quadratic Classifier) have been tested under a set of experiments using an available database. The results have demonstrated that the Quadratic Classifier provides errors less than Classifier. However, the generalization problems are less using Linear Classifier.

Using MFCC-based enlargement of the training set, the system has a number of patterns appropriate, and it is possible train to the system correctly. With this enlargement of the database, it is possible to reduce the error probability in emotion recognition near 8%, which is a considerable improvement in the performance.

## ACKNOWLEDGEMENTS

This work has been funded by the Spanish Ministry of Education and Science (TEC2012-38142-C04-02), by the Spanish Ministry of Defense (DN8644-ATREC) and by the University of Alcala under project CCG2013/EXP-074.

# AUTHORS

**Inma Mohino-Herranz**

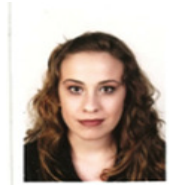

Year in which an academic degree was awarded: Telecommunication Engineer, Alcalá University, 2010. PhD student about Information and Communication Technologies. Area of research: Signal Processing.

**Roberto Gil-Pita**

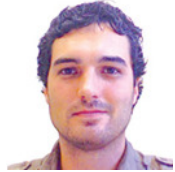

Year in which an academic degree was awarded: Telecommununication Engineer, Alcalá University, 2001. Possition: Associate Professor. Polytechnic School in the Department of Signal Theory and Communications. Some of his research interest include, audio, speech, image, biological signals.

**Sagrario Alonso-Díaz**

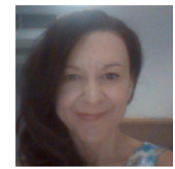

PhD Psicologist. Researcher in the Human Factors Unit. Technological Institue "La Marañosa" –MoD.

**Manuel Rosa-Zurera,**

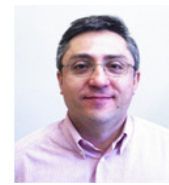

Year in which an academic degree was awarded: Telecomunication Engineer, Polythecnic University of Madrid, 1995. Possition: Full professor and dean of Polytechnic School. University of Alcalá. His areas of interest are audio, radar, speech source separation.